\documentclass{article}

\usepackage[preprint, nonatbib]{nips_2018}
\usepackage{amsmath}
\usepackage{makecell}

\usepackage{float}

\usepackage[square,numbers]{natbib}
\usepackage[utf8]{inputenc} 
\usepackage[T1]{fontenc}    
\usepackage{hyperref}       
\usepackage{url}            
\usepackage{booktabs}       
\usepackage{amsfonts}       
\usepackage{nicefrac}       
\usepackage{microtype}      
\usepackage{graphicx}
\usepackage{tabularx}
\usepackage{hyperref}

\title{StairNetV3: Depth-aware Stair Modeling using Deep Learning}

\author{
 Chen Wang \\
  \texttt{School of Automation Science and Electrical Engineering}\\
  \texttt{Beihang University, Beijing, 100191, China}\\
  \texttt{venus@buaa.edu.cn} \\
\And
 Zhongcai Pei \\
  \texttt{School of Automation Science and Electrical Engineering}\\
  \texttt{Beihang University, Beijing, 100191, China}\\
  \texttt{peizc@buaa.edu.cn} \\
\And
 Shuang Qiu \\
  \texttt{School of Automation Science and Electrical Engineering}\\
  \texttt{Beihang University, Beijing, 100191, China}\\
  \texttt{zb2003108@buaa.edu.cn} \\
\And
 Yachun Wang \\
  \texttt{School of Automation Science and Electrical Engineering}\\
  \texttt{Beihang University, Beijing, 100191, China}\\
  \texttt{yachun@buaa.edu.cn} \\
\And
 Zhiyong Tang \\
  \texttt{School of Automation Science and Electrical Engineering}\\
  \texttt{Beihang University, Beijing, 100191, China}\\
  \texttt{zyt\_76@buaa.edu.cn} \\
}

\begin{document}
\maketitle

\begin{abstract}
Vision-based stair perception can help autonomous mobile robots deal with the challenge of climbing stairs, especially in unfamiliar environments. To address the problem that current monocular vision methods are difficult to model stairs accurately without depth information, this paper proposes a depth-aware stair modeling method for monocular vision. Specifically, we take the extraction of stair geometric features and the prediction of depth images as joint tasks in a convolutional neural network (CNN), with the designed information propagation architecture, we can achieve effective supervision for stair geometric feature learning by depth information. In addition, to complete the stair modeling, we take the convex lines, concave lines, tread surfaces and riser surfaces as stair geometric features and apply Gaussian kernels to enable the network to predict contextual information within the stair lines. Combined with the depth information obtained by depth sensors, we propose a stair point cloud reconstruction method that can quickly get point clouds belonging to the stair step surfaces. Experiments on our dataset show that our method has a significant improvement over the previous best monocular vision method, with an intersection over union (IOU) increase of 3.4 $ \% $, and the lightweight version has a fast detection speed and can meet the requirements of most real-time applications. Our dataset is available at \href{https://data.mendeley.com/datasets/6kffmjt7g2/1}{https://data.mendeley.com/datasets/6kffmjt7g2/1}

\end{abstract}

\section{Introduction}

In recent years, with the development of autonomous humanoid robots, the perception and feature representation of the surrounding environments, especially various obstacles, have gradually become a research hotspot. Stairs are common indoor and outdoor building structures in urban environments, which are difficult for robots to pass through. Because stairs are artificially constructed building structures with obvious geometric features, most stair detection methods rely on extracting some stair geometric features. For example, line-based extraction methods \cite{bib2, bib3, bib1} abstract the stair geometric features as a set of lines continuously distributed in an image, and potential stair lines are extracted in RGB or depth images through Canny edge detection \cite{bib4}, Sobel filtering, Hough transform \cite{bib5}, etc. Plane-based extraction methods \cite{bib6, bib7, bib8} abstract the stair geometric features as a set of parallel planes continuously distributed in space, and potential stair surfaces in point cloud data are extracted through algorithms such as random sample consensus (RANSAC) \cite{bib9} and supervoxel clustering \cite{bib10}.

The works mentioned above extract stair features using artificially designed feature extraction rules, and these methods may fail to some extent in complex scenes. To deal with complex and challenging scenes, machine learning techniques have been applied to stair detection. For visually impaired people, it is only necessary to determine whether the environment contains stairs. support vector machines (SVM) \cite{bib11} can be used to determine whether there are stairs in an image \cite{bib12, bib13}. For autonomous robots, the stair modeling is needed to pass through stairs successfully. Some works first locate the region of interest (ROI) containing stairs through object detection algorithms such as you only look once (YOLO) \cite{bib14}, and then extract stair features within the ROI through traditional image processing algorithms \cite{bib15}. These methods with two steps often have poor real-time performance. To achieve end-to-end stair feature extraction, StairNet \cite{bib16} proposes a novel stair feature representation method which is conducive to neural network learning, and uses a simple fully convolutional network to complete the extraction of stair line features. StairNet does not rely on artificially designed rules and shows strong robustness and adaptability to complex scenes, and end-to-end detection has good real-time performance. To overcome the poor performance of StairNet in scenes with extreme lighting conditions and scenes with extremely fuzzy visual clues, StairNetV2 \cite{bib17} adds depth information input to the network to explore the complementary relationship between RGB images and depth images. However, StairNetV2 relies on dense depth images generated by binocular vision. Without binocular sensors, StairNetV2 will not be applicable and its applicability is inferior to monocular vision methods.

\begin{figure}[!htb]
\includegraphics[width=0.85\linewidth]{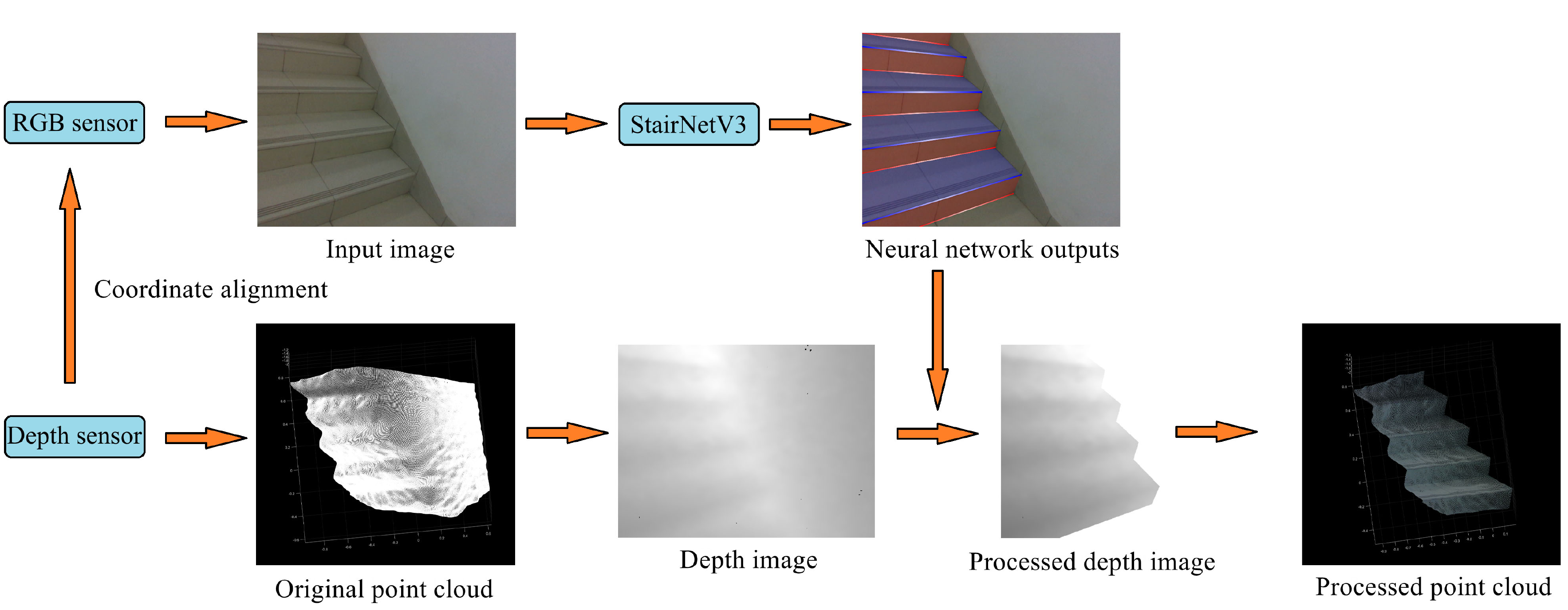}
\caption{The overall flow of the proposed method. After the RGB image captured by the monocular camera is processed by StairNetV3, the stair modeling results are obtained, including convex lines (blue lines in the figure), concave lines (red lines in the figure), tread surfaces (blue masks in the figure), riser surfaces  (red masks in the figure) and stair line endpoints. After the depth sensor (depth cameras, lidars, etc.) obtains the environmental point cloud and the corresponding depth matrix, the segmentation outputs of StairNetV3 are applied to the depth matrix, combined with the RGB image and camera intrinsics, the point clouds of the stair step surfaces are obtained through point cloud reconstruction.}
\label{fig1}
\end{figure}

To solve the above problems, we propose depth-aware StairNetV3 to learn depth features through depth supervision under the condition of monocular vision, which can retain the wide applicability of StairNet while achieve detection performance close to StairNetV2. In addition, we improve the feature representation  method of the StairNet series by adding Gaussian kernels, making the network can perceive stair line endpoints. To achieve complete modeling of stair building structures, we add a semantic segmentation branch to extract stair step surfaces while extracting stair lines. Combined with point cloud data obtained from depth sensors, stair point cloud segmentation can be achieved at a very fast speed. The overall flow of our method is shown in Figure~\ref{fig1}.

The remainder of this paper is organized as follows: Section~\ref{sec2} summarizes the related works on stair detection, Section~\ref{sec3} discusses the details of our methods, including the improved stair feature representation method, the network architecture, the stair point cloud reconstruction method and the loss function. Section~\ref{sec4} discusses the experiments conducted, including dataset and metrics, implementation details, ablation experiments, performance experiments and comparison experiments. Section~\ref{sec5} summarizes the whole paper and gives an outlook for future work.

\section{Related Work}
\label{sec2}

Stairs are a common architectural structure, with different materials, structures, and design styles. Stair detection in different environments is a challenging task. Some works are dedicated to detecting stairs in monocular vision. For example, reference \cite{bib18} applies Gabor filters on grayscale RGB images to extract stair edges and judges whether the scene contains stairs through the projection-histogram of edge pixel values in the horizontal and vertical directions. Reference \cite{bib19} regards the stair line as a periodic signal in the spatial domain. The input RGB image is processed through Canny edge detection to obtain a binary image, which is then processed through 2D fast Fourier transform to the frequency domain, resulting in an image that almost only contains stair edges. Reference \cite{bib20} regards the stair line endpoints as the intersection of three line segments and searches for such points in the edge image after Gabor filtering to extract potential stair lines. With the development of machine learning and deep learning, stair detection has gradually shifted from traditional image processing dominated methods to deep learning dominated methods. For examples, reference \cite{bib21} proposes a lightweight Look-Behind fully convolutional network to determine whether there are stairs in an image. Reference \cite{bib22} applies Adaboost classifier and Haar features to detect stairs in an image, and uses ground plane estimation and temporal consistency validation to reduce the number of false positives. Reference \cite{bib23} first locates the box containing stairs through YOLOv5s \cite{bib24}, and then applies U-Net \cite{bib25} with ResNet-34 \cite{bib26} backbone to segment stair lines within the box. This research achieves pixel-level location of stair lines through deep learning methods, but due to the multi-stage execution of the methods, the whole real-time performance is poor. Reference \cite{bib16} achieves end-to-end detection of stair lines through a fully convolutional neural network by designing a novel stair feature representation method.

In recent years, with the decline in the prices of binocular sensors and lidars, more and more research has been devoted to detecting stairs through binocular vision and stereo vision. Most binocular vision detection methods are based on monocular detection methods, using depth features from depth images to remove interfering textures. For example, references \cite{bib27, bib28} first detect stair edges in RGB images through traditional image processing methods, and then extract corresponding one-dimensional depth features in depth images to distinguish between stairs and pedestrian crosswalks. Reference \cite{bib29} applies Sobel filtering and Hough transform to detect potential lines, and then extracts a 480-dimensional feature vector in the corresponding depth image and feeds the vector into SVM to determine whether the scene is ascending stairs, descending stairs or other situations. These methods combine features from different modalities through artificially designed rules, and their adaptability in complex scenes is difficult to guarantee. To explore the potential complementary relationship between RGB images and depth images, StairNetV2 \cite{bib17} adds selective module to the network to obtain effective multimodal combination automatically through learning, greatly improving the detection reliability in complex scenes, especially at night. The main idea of stereo vision-based methods is to extract point clouds of stair step surfaces in 3D space. For example, reference \cite{bib30} first uses RANSAC to search for planes, then uses principal component analysis (PCA) to estimate the normal and surface curvature of the planes and connect adjacent planes. After distinguishing obstacles in the scene based on the proportion of inliers, the riser surfaces and tread surfaces are determined by the angle between the normal of candidate planes and the normal of the ground plane. Reference \cite{bib31} uses region growing clustering to cluster point clouds obtained from depth cameras, effectively distinguishing stair step surfaces and walls, and avoiding interference from obstacles through histograms of the inlier number on planes at different heights. Reference \cite{bib32} uses PointNet \cite{bib33} to process point clouds obtained from depth cameras and classifies the scene into ascending stairs, descending stairs or other situations.

\section{Methods}
\label{sec3}
In this section, we describe the methods in detail, including the feature representation method of stairs, the detailed architecture of the network, the post-processing algorithms for network outputs, and the loss function.

\subsection{Stair feature representation}
\label{sec3_1}
Different from the approach of only focusing on line extraction in StairNet and StairNetV2, StairNetV3 is dedicated to achieving complete modeling of stair structures. We use a simple neural network to simultaneously conduct line extraction and surface extraction, distinguishing between convex lines/concave lines and riser surfaces/tread surfaces. For stair lines, endpoint information is crucial in the post-processing of network results. We hope that the network can directly learn the stair line endpoints, so we introduce Gaussian kernels.

The input of the network is a full-color RGB image, denoted as $ I \in [0, 255]^{H \times W \times 3} $, and its height and width are $ H $ and $ W $, respectively. Our goal is to obtain a heatmap that reflects the position of stair lines, denoted as $ O \in [0, 1]^{\frac{H}{S_H} \times \frac{W}{S_W} \times 1} $, where $ S_H $ and $ S_W $ are the output strides in the width and height directions, respectively, determining the heatmap size. To express the endpoint information of stair lines, we hope that the response of stair line endpoints in the heatmap is 1, while the response of the rest of the stair lines decreases as the distance from the endpoint increases. Therefore, we apply Gaussian kernels to determine the value of each pixel in the heatmap, as shown in equation~\ref{eq1}. 

\begin{equation}
	Y=
	\begin{cases}
	
	e^{-\frac{(x-p_x)^2+(y-p_y)^2}{2\sigma^2}}, & if (x, y) \in L \\
	0,& \text{otherwise}
	
	\end{cases}\label{eq1}
\end{equation}

\begin{figure}[!htb]
\includegraphics[width=0.82\linewidth]{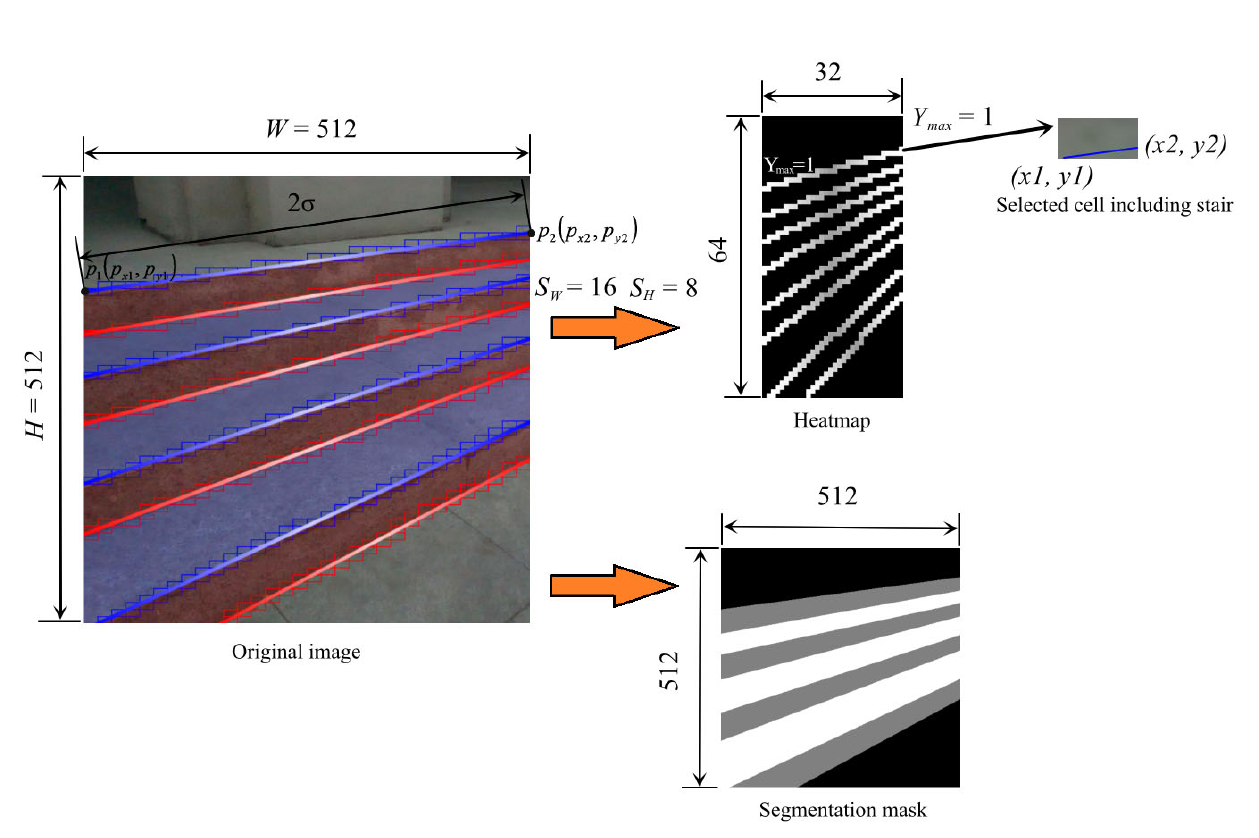}
\caption{Illustration of stair feature representation. The line features include convex lines (blue lines in the original image) and concave lines (red lines in the original image), and the surface features include tread surfaces (blue masks in the original image) and riser surfaces (red masks in the original image). To obtain the stair line endpoints, Gaussian kernels are applied so that the two stair line endpoints obtain the maximum response of 1 in the heatmap, and the stair line midpoint obtains the minimum response, approximately 0.6. The color of each stair line in the original image gradually fades from both ends to the middle, reflecting the working mode of the Gaussian kernels. For cells containing stair lines, the normalized coordinates $ (x1,y1) $ and $ (x2,y2) $ of the stair line relative to the upper left corner of the cell are regressed. The stair step surfaces are predicted through an output segmentation mask with the same size as the original image.}
\label{fig2}
\end{figure}

Where $ (x, y) $ represents the pixel coordinates in the original image, $ Y $ represents the pixel value at the corresponding pixel coordinates $ (\biggl\lfloor\frac{x}{S_W}\biggr\rfloor, \biggl\lfloor\frac{y}{S_H}\biggr\rfloor) $ in the heatmap, $ (p_x, p_y) $ represents the pixel coordinates of the closer endpoint of stair line $ L $ to $ (x,y) $ in the original image, and $ \sigma $ is the standard deviation and can be adjusted according to demand \cite{bib34}. According to equation~\ref{eq1}, it can be seen that when $ (x,y) $ is the midpoint of stair line $ L $, $ Y $ takes the minimum value. Since the threshold for distinguishing positive and negative samples is usually set to 0.5, we hope that the $ Y $ value at the midpoint is greater than 0.5. Therefore, let $ \sigma=\frac{1}{2}||L||_2 = \frac{1}{2}\sqrt{(p_{x1}-p_{x2})^2+(p_{y1}-p_{y2})^2} $, where $ (p_{x1}, p_{y1}) $ and $ (p_{x2}, p_{y2}) $ represents the left and right endpoints of stair line $ L $ in the original image, respectively. So when $ x=\frac{p_{x1}+p_{x2}}{2} $ and $ y=\frac{p_{y1}+p_{y2}}{2} $, $ 
Y_{min}=e^{-\frac{1}{2}}\approx0.6 $. For the prediction of stair step surfaces, we add a branch to the network for generating semantic segmentation masks. The branch has four upsampling operations to obtain tensors that match the input image size, and the segmentation results have three classifications, including riser, tread, and background. The above process is shown in Figure~\ref{fig2}.

\subsection{Network architecture}
\label{sec3_2}
\begin{figure}[!htb]
\includegraphics[width=\linewidth]{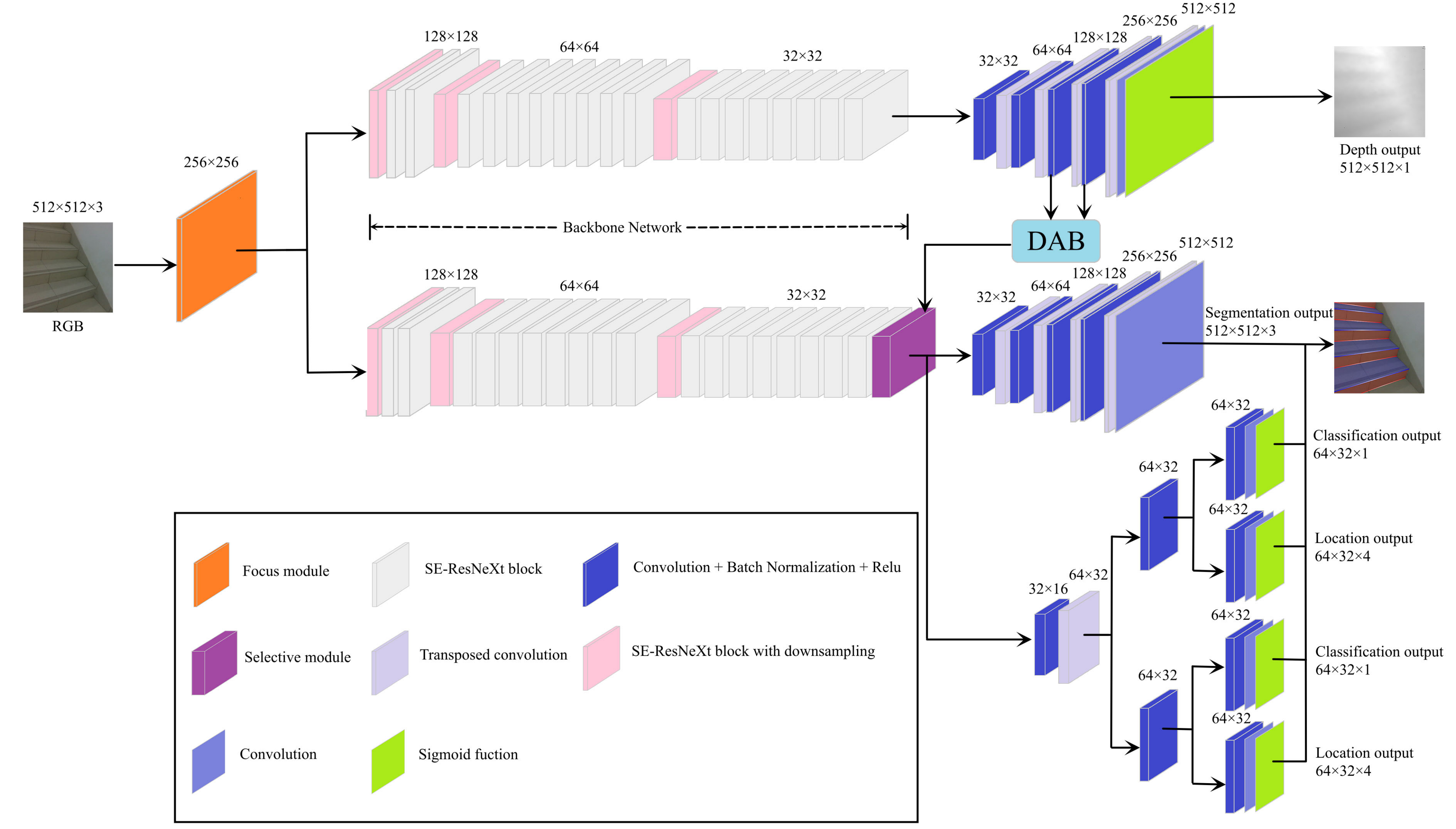}
\caption{Network architecture. The network contains two parallel backbones, and the upper backbone is used to output depth images and the lower backbone is used to output stair modeling information. The depth features from different feature layers are fused and passed to the stair modeling backbone through the DAB, and the feature fusion is achieved through a selective module, enabling direct supervision of the stair modeling results by the depth-aware information.}
\label{fig3}
\end{figure}

Different from the network architecture with two input branches in StairNetV2, StairNetV3 only has an input branch of RGB image, and the branch of depth image is set in the output branches. To achieve effective depth-awareness, the network contains two parallel backbones, which are stacked with several squeeze-and-excitation (SE)-ResNeXt \cite{bib35, bib36} blocks with dilated convolution. Both backbones follow an encoder-decoder structure, where one backbone is used for depth-awareness and the other backbone is used for obtaining stair modeling information. The interaction between the two backbones includes two types. Through the shared input branch, the stair modeling backbone get the implicit supervision from depth information by the way of gradient back propagation. In the decoder of the depth-awareness backbone, a depth-aware block (DAB) is applied to aggregate depth information from different feature layers and directly pass it to the encoder of the stair modeling backbone, and the feature fusion is achieved through a selective module \cite{bib17} to directly supervise the learning of stair modeling. The network output includes 6 branches. The depth-awareness backbone outputs an estimated depth image with a size of 512 $ \times $ 512 $ \times $ 1. For the stair modeling backbone, to make the network classification results more robust, we decouple the output branches of the two types of stair lines, so there are a total of 4 branches, including 2 heatmap branches with a size of 64 $ \times $ 32 $ \times $ 1 and 2 location branches with a size of 64 $ \times $ 32 $ \times $ 4. The heatmap branch outputs the confidence that the cell contains stair line and the location branch outputs the normalized coordinates of the contained stair lines. In addition, the stair modeling backbone also includes a semantic segmentation branch with a size of 512 $ \times $ 512 $ \times $ 3 for classifying stair step surfaces. The network architecture is shown in Figure~\ref{fig3}.

\subsubsection*{Implicit depth supervision}

In StairNetV3, the depth-aware backbone and the stair modeling backbone share parameters through the common focus \cite{bib24} module. The focus module is essentially a tensor slicing operation, and the feature maps obtained by slicing and stacking are adjusted in channel number and fused across channels through point convolution. Specifically, given an input image  
$ I \in [0, 255]^{H \times W \times 3} $ with height and width of $ H $ and $ W $ respectively. After the focus module, we can get $ f_{\theta}(I^{'}) $, where $ f_\theta $ represents the weights of point convolution and $ I^{'} \in [-1, 1]^{\frac{H}{2} \times \frac{W}{2} \times 12} $, then the entire learning process can be considered as the following optimization problem, as show in equation~\ref{eq2}.

\begin{equation}
    \mathop{\text{min}}\limits_\theta N_D(f_{\theta}(I^{'});D^{*})+\mathop{\text{min}}\limits_\theta N_M(f_{\theta}(I^{'});M^{*})
\label{eq2}
\end{equation}

Where, $ N_D $ and $ N_M $ represent data fitting terms, $ D^{*} $ and $  M^{*} $ represent the ground-truth of depth image and the ground-truth of stair modeling respectively. Through sharing the weights $ f_{\theta} $ of point convolution in focus module and gradient back propagation, the point convolution can generate weights that are beneficial for obtaining depth images, and the depth images are enlightening for the stair modeling, so the learning process of stair modeling is optimized in an implicit depth supervision manner.

\subsubsection*{Depth-aware block}

\begin{figure}[!htb]
\includegraphics[width=\linewidth]{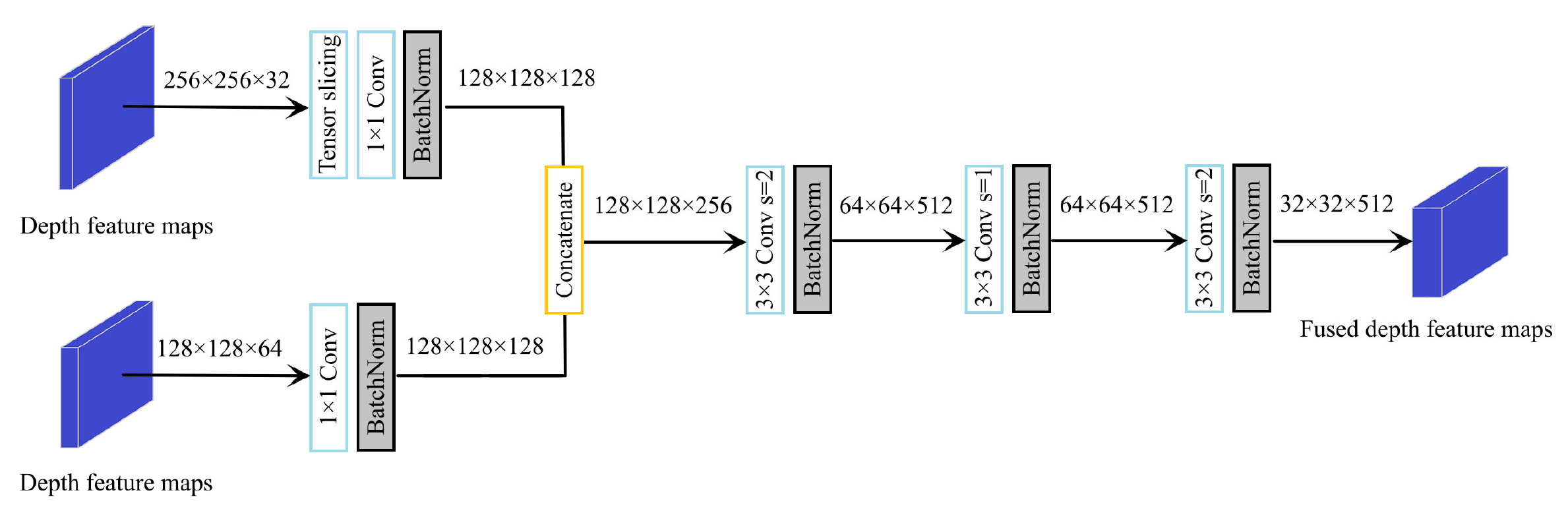}
\caption{The structure of DAB. The DAB first fuses the feature maps obtained from the second and third upsampling layers of the decoder in the depth backbone through a 1 $ \times $ 1 convolution and tensor slicing operation. The fused feature maps are then processed through three 3 $ \times $ 3 convolutions to obtain fused feature maps with a size of 32 $ \times $ 32 $ \times $ 512. To ensure that the selective module can obtain the original depth information as much as possible, there is no activation function in DAB, and a 3 $ \times $ 3 convolution is applied in the downsampling to avoid feature loss.}
\label{fig4}
\end{figure}

In StairNetV2, the selective module makes full use of the complementary relationship between depth images and RGB images, improving the perceptual ability when visual clues are fuzzy. The optimization goal of the depth backbone in StairNetV3 is to regress the depth image aligned to the input RGB image. Therefore, we take the feature maps obtained after the second and third upsampling in the decoder of the depth backbone, and fuse them through the DAB. Then the fused depth feature maps are sent to the selective module to obtain direct depth supervision for the learning of stair modeling backbone, further improving the  stair modeling performance. The DAB is shown in Figure~\ref{fig4}.

\subsection{Post-processing algorithms}
\label{sec3_3}

\subsubsection*{Line segments linking method}

For stair detection algorithms based on line extraction, the outputs are often a set of discrete stair lines. To obtain the stair line equations, these discrete line segments need to be linked and fitted. The Line extraction method based on traditional image processing often use edge connection algorithms to connect adjacent line segments \cite{bib12, bib13, bib20}. For the structured stair line information output by convolutional network, StairNetV2 proposes a line clustering method based on least squares \cite{bib17}, which can quickly complete clustering using adjacent information and geometric information between stair line segments. However, repeated least squares calculations mean that the algorithm still requires some processing time in real-time inference. To make the post-processing of line linking faster, we assign higher-level semantic information to each cell containing stair lines through Gaussian kernels. Each cell can not only obtain whether it contains a stair line, but also obtain its relative distance from the stair endpoint (which can be understood as the confidence of containing a stair line endpoint in the cell). Therefore, the line segments linking method can directly obtain the cells containing stair line endpoints which are more critical to the line segments linking, and the detailed steps are as follows.

\begin{enumerate}
\item The cells with confidence greater than a certain threshold (set to 0.75 in the research) are selected and sorted according to their confidence scores.
\item After the cells with high confidence scores are selected, the top 50 cells with highest confidence scores are selected and the adjacent cells are grouped together.
\item Groups with only one cell are removed.
\item For each groups, the stair lines in the cells are fitted through least-squares method to get the line equation $ y=kx+b $.
\item For each two groups, their line equations are checked. If the equations intersect in the range of [0, 512], or if their left endpoints are close or their right endpoints are close, then the two groups are merged into a new group. After that, the slope-intercept equations of the lines are recalculated.
\end{enumerate}

\begin{figure}[!htb]
\includegraphics[width=\linewidth]{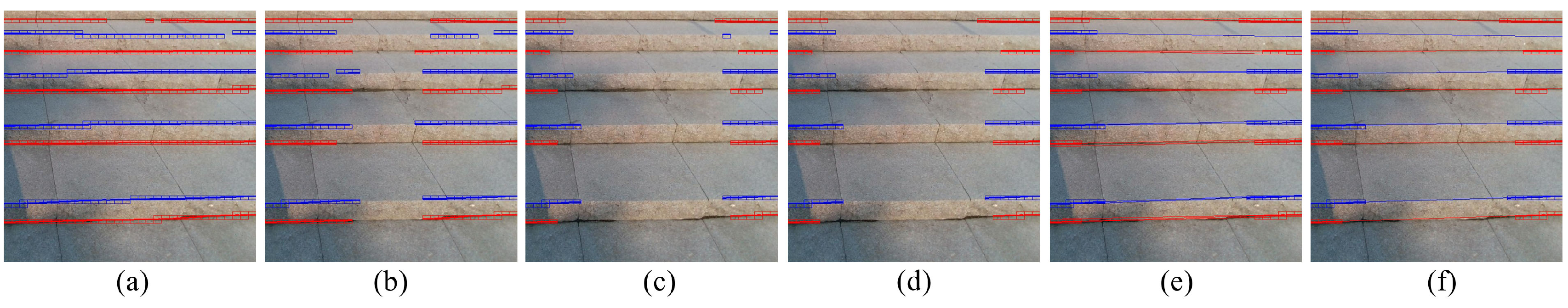}
\caption{Illustration of line linking method. Figure (a) shows the original stair line detection results output by StairNetV3; Figure (b) shows the results after filtering with a threshold of 0.75; Figure (c) shows the top 50 cells with the highest confidence scores and the stair lines they contain; Figure (d) shows the results after filtering isolated cells; Figure (e) shows the initial fitted stair lines through the least squares method; Figure (f) shows the final fitted results after the merging of groups.}
\label{fig5}
\end{figure}

Overall, the line segments linking method can obtain the equations of stair lines through information from only 50 cells and 2 rounds of least squares calculation, as shown in Figure~\ref{fig5}. 

\subsubsection*{Stair point cloud reconstruction method}

After obtaining the stair line equations, To achieve a complete stair modeling, it is necessary to extract the point clouds of the stair step surfaces. Most plane-based methods apply point cloud segmentation algorithms in point clouds to obtain stair step surfaces. Due to the large number of point cloud data, the real-time performance of these methods is often poor. To achieve fast stair surface segmentation, we transform the segmentation operation from three-dimensional space to two-dimensional images. Through the semantic segmentation branch of the network, we directly obtain the stair step surfaces in the images and apply the segmentation results to the depth matrix. Then, combined with the camera intrinsic matrix, we obtain the point clouds of the stair step surfaces in a point cloud reconstruction manner. Specifically, given a point cloud data $ P \in R^{X,Y,Z} $ from a depth sensor, the distance $ Z \in [dmin, dmax]^{H \times W} $ in the forward direction can be obtained, namely the depth matrix, where $ H $ and $ W $ are the height and width of the aligned RGB image. The output of the network segmentation branch is denoted as $ S \in [0,1]^{H \times W \times 3} $, and the third dimension of $ S $ predicts the one-hot encoding $ (c_1, c_2, c_3) $ for three classifications, then the maximum value is retained and set to 1 and other values are set to 0. After the above transformation, each element in $ S $ can be described as the following equation~\ref{eq3}

\begin{equation}
	S_{h,w,i}^*=
	\begin{cases}
	
	1, & if S_{h,w,i} \in c_i \\
	0, & \text{otherwise}
	
	\end{cases}\label{eq3}
\end{equation}

Where, $ h \in [0,H) $, $ w \in [0,W) $, $ i \in [0,3) $, for each classification $ i $, the corresponding 0-1 matrix $ S_i^* $ is obtained, and through dot product, the corresponding depth matrix $ Z_i^*=Z*S_i^* $ is obtained. For each element $ Z_{h,w,i}^* $ in $ Z_i^* $, the corresponding coordinates in camera coordinate system can be obtained trough the flowing equation, as shown in equation~\ref{eq4}.

\begin{equation}
	P_{h,w,i}^*=
	\begin{cases}
	
	K^{-1}Z_{h,w,i}^*p_{h,w}, & if Z_{h,w,i}^* \neq 0 \\
	(0,0,0), & \text{otherwise}
	
	\end{cases}\label{eq4}
\end{equation}

\begin{figure}[!htb]
\includegraphics[width=\linewidth]{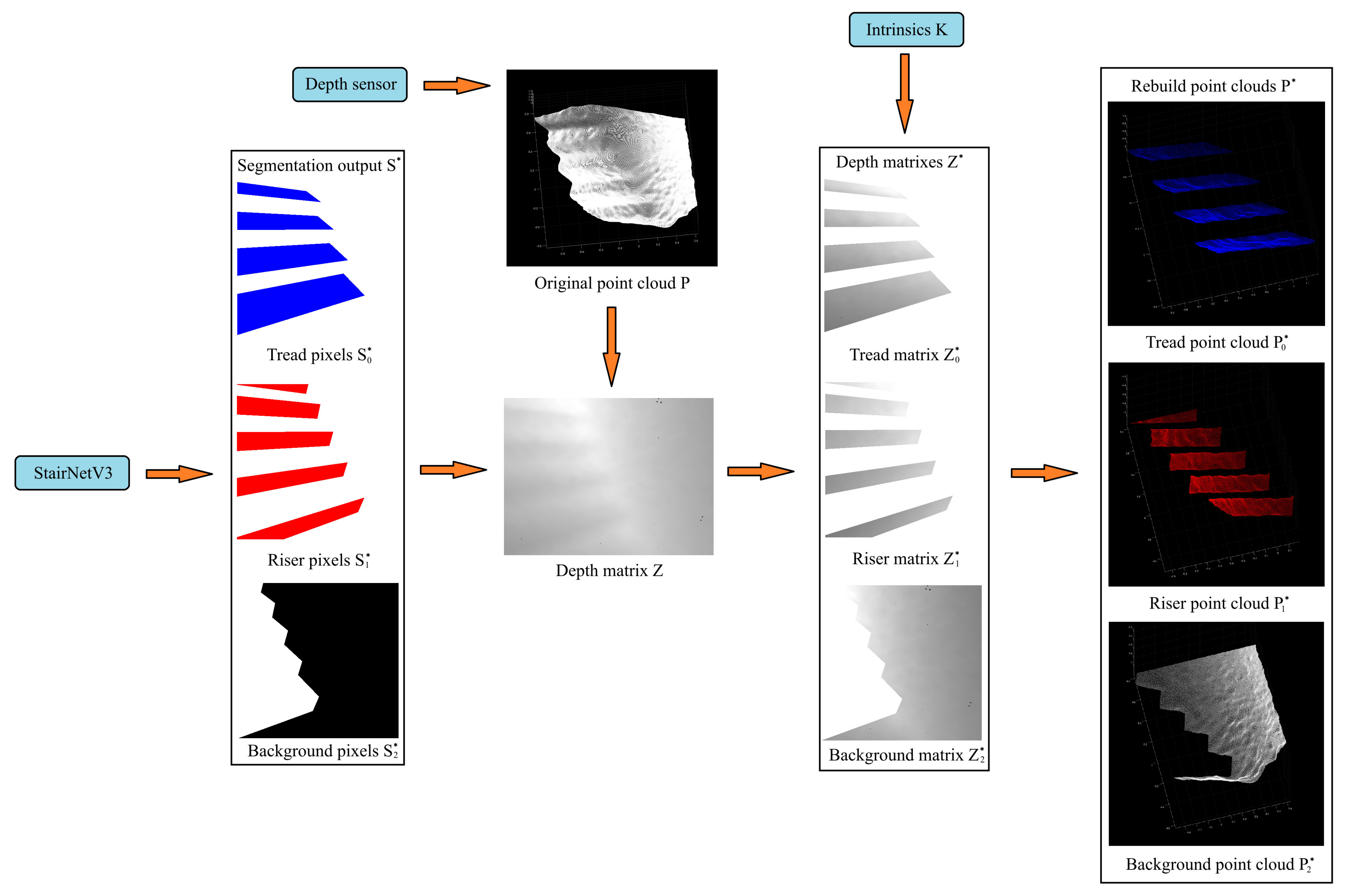}
\caption{Illustration of stair point cloud reconstruction. After obtaining the point cloud data from depth sensor, the depth matrix $ Z $ is generated. After obtaining the segmentation results of StairNetV3, the 0-1 matrix $ S_i^* $ can be obtained for each classification. Trough dot product, the depth matrix $ Z_i^* $ for each classification is obtained. Finally, combined with the camera intrinsics $ K $, the point cloud $ P_i^* $ corresponding to each classification is obtained through point cloud reconstruction.}
\label{fig6}
\end{figure}

Where $ P_{h,w,i}^* $ are the coordinates in camera coordinate system, $ K $ is the camera intrinsic matrix, $ p_{h,w} $ are the coordinates in pixel coordinate system, and $ h $, $ w $ and $ i $ have the same meaning as in equation (3). After the above process, we can obtain the point cloud $ P_i^* $ belonging to the classification $ i $, as shown in Figure~\ref{fig6}.

\subsection{Loss function}
\label{sec3_4}
StairNetV3 has multiple output branches, which can be divided into branches for classification and branches for location. For the classification task, we use the Binary Cross Entropy (BCE) loss function with a sigmoid function. For the regression task, we use the Mean Squared Error (MSE) loss function. For the stair line extraction task, we regard it as a regression problem, and the loss function is described as equation~\ref{eq5}:

\begin{equation}
    L_l=\frac{1}{MN}\sum_{i}^{M}\sum_{j}^{N}(\alpha_{1}h_{ij}L_{MSE}(p_{ij},p_{ij}^*)+\alpha_{2}(1-h_{ij})L_{MSE}(p_{ij},p_{ij}^*)+h_{ij}L_{MSE}(t_{ij},t_{ij}^*))
\label{eq5}
\end{equation}

Where, $ M $ and $ N $ represents the height and width of the output feature maps, namely, 64 and 32, and $ i $ and $ j $ represents the position of the cell in the feature maps. $ L_{MSE} $ represents MSE loss function, $ p_{ij} $ represents the confidence predicted by each cell, and $ p_{ij}^* $ is the corresponding ground truth. $ h_{ij} $ represents the prediction probability of containing stair line and its values are 0 and 1, which represent containing stair line and without stair line, respectively.  $  t_{ij} $ represents the normalized coordinates of stair line endpoints in each cell, and $ t_{ij}^* $ is the corresponding ground-truth. For the confidence loss, we give different weights to positive and negative samples to adjust the imbalance between positive samples and negative samples, and $ \alpha_1 $ and $ \alpha_2 $ represents the weight coefficients of positive samples and negative samples, respectively. In our research, $ \alpha_1 $=15, $ \alpha_2 $=5.

For location loss, we inherit the loss function with dynamic weights \cite{bib17} from StairNetV2, as shown in equation~\ref{eq6}.

\begin{equation}
    L_{MSE}(t_{ij},t_{ij}^*)=\lambda_{1}L_{MSE}(x_{ij},x_{ij}^*)+\lambda_{2}L_{MSE}(y_{ij},y_{ij}^*)
\label{eq6}
\end{equation}

Where, $ x_{ij} $ represents the predicted normalized coordinates in the horizontal direction, and $ x_{ij}^* $ is the corresponding ground truth. $ y_{ij} $ represents the predicted normalized coordinates in the vertical direction, and $ y_{ij}^* $ is the corresponding ground truth. $ \lambda_1 $ and $ \lambda_2 $ represents the corresponding dynamic weights, respectively. In initial epoch, $ \lambda_1 $=$\lambda_2 $=10, After each epoch, the dynamic weights are adjusted automatically according to the evaluation indexes. The rest parameters have the same meanings as equation~\ref{eq5}.

For the semantic segmentation task of the stair step surfaces, its essence is a classification task. We apply BEC loss and its loss function is shown in equation~\ref{eq7}:

\begin{equation}
    L_s=\frac{1}{N^{2}C}\sum_{i}^{N}\sum_{j}^{N}\sum_{k}^{C}L_{BCE}(m_{ij},m_{ij}^*)
\label{eq7}
\end{equation}

Where, $ N $ represents the width or height of the output feature maps, namely, 512. $ C $ represents the number of classifications, which is 3, $ i $ and $ j $ represents the position of the pixel in the feature maps, and $ k $ represents the classification.  $ L_{BCE} $ represents the BCE loss function. $ m_{ij} $ represents the prediction probability of a pixel belonging to a certain classification, and $ m_{ij}^* $ is the corresponding ground-truth. 

For the depth estimation task, it is essentially a regression task, and we still use the MSE loss, and the loss function is shown in equation~\ref{eq8}:

\begin{equation}
    L_d=\psi\frac{1}{N^{2}}\sum_{i}^{N}\sum_{j}^{N}L_{MSE}(d_{ij},d_{ij}^*)
\label{eq8}
\end{equation}

Where $ N $, $ i $ and $ j $ have the same meanings as in equation~\ref{eq7}, $ L_{MSE} $ represents the MSE loss function, $ d_{ij} $ represents the predicted normalized pixel value, and $ d_{ij}^* $ is its corresponding ground-truth. $ \psi $ is a weight coefficient, and in this study, $ \psi $=10. Overall, the loss function $ L $ of StairNetV3 can be represented as equation~\ref{eq9}:

\begin{equation}
    L=L_l^b+L_l^r+L_s+L_d
\label{eq9}
\end{equation}

Where, $ L_l^b $ and $ L_l^r $ represent the loss of convex lines and the loss of concave lines respectively.

\section{Experiments}
\label{sec4}

\subsection{Dataset and metrics}
\label{sec4_1}

The data used to evaluate our method all come from the Stair dataset with depth maps \cite{bib37}. Based on the dataset, we remove the images taken under no light source conditions, and finally retain 2276 RGB-D image pairs as the training set and 556 RGB-D image pairs as the validation set. To evaluate the performance of point cloud reconstruction, we use the Realsense D435i depth camera \cite{bib38} to collect 154 RGB-D image pairs with a resolution of 640 $ \times $ 480 as the test set. During the collection, we record the pose of camera and the linear transformation relationship of depth image generation for point cloud reconstruction. To implement the semantic segmentation of stair step surfaces, we make pixel-level segmentation labels for the training set and the validation set, and the classifications include stair riser surfaces, stair tread surfaces and background. The organized dataset is named as the RGB-D stair dataset \cite{bib39}.

To accurately evaluate our method, we apply several normal evaluation metrics. For stair lines, we use precision, recall, and IOU with a confidence level of 0.5 for evaluation. This is consistent with the evaluation methods of StairNet and StairNetV2. For stair step surfaces, we use pixel accuracy (PA) and mean pixel accuracy (MPA) for evaluation.

\subsection{Implementation details}
\label{sec4_2}

We implement StairNetV3 with the PyTorch framework on a workstation that has an i9 13900 CPU and an RTX 3090 GPU. During training, we use Adam optimizer \cite{bib40} with weight decay of $ 10^{-6} $ to optimize the network. The learning rate is initialized as 2.5 $ \times  10^{-4} $ and halved every 50 epochs. The batch size is set to 4. The input images have the size of 512 $ \times $ 512, and the random horizontal flip with a probability of 0.5 is applied for data augmentation.

\subsection{Ablation study}
\label{sec4_3}

To verify the role of the DAB and explore its optimal combination with implicit deep supervision, we conduct several ablation experiments. We take a model without DAB as the baseline, and other models place the DAB after the second downsampling, the third downsampling and the fourth downsampling for comparison. All ablation experiments follow the same settings and procedures, and the results are shown in Table~\ref{tab1}.

\begin{table}[!htb]
  \caption{The ablation experimental results of DAB.}
  \label{tab1}
  \centering
    \resizebox{.9\textwidth}{!}{
    \begin{tabularx}{\textwidth}{>{\centering}cccccc}
    \toprule
    \textbf{The position of DAB} & \textbf{Precision ($ \% $)} & \textbf{Recall ($ \% $)} & \textbf{IOU ($ \% $)} & \textbf{PA ($ \% $)} & \textbf{MPA ($ \% $)} \\
    \midrule
    Without DAB                   & 77.00 & 81.27 & 65.40 & 96.03 & 81.02\\
    \midrule
    After the second downsampling & 76.99 & 81.22 & 65.36 & 96.09 & 81.08\\
    \midrule
    After the third downsampling  & 77.16 & 81.86 & 65.89 & 96.31 & 81.25\\
    \midrule
    After the fourth downsampling & 77.25 & 81.80 & 65.92 & 96.30 & 81.27\\
    \bottomrule
    \end{tabularx}
    }
\end{table}

Table~\ref{tab1} shows that the position of the DAB can impact on network performance. When the DAB is placed after the second downsampling, the model performance is even comparable to that without DAB. When the DAB is placed after the third downsampling and fourth downsampling, the model performance is significantly improved. This is mainly because the DAB placed after the third and fourth downsampling can achieve feature fusion at different levels. The DAB contains deep features of the decoder, and the depth features of DAB are closer to the features of original depth image. These features correspond to the shallow features of encoder. Through the feature fusion of deep features and shallow features in the encoder, we realize a method of multi-level feature map fusion. Compared with the fusion of shallow features after the second downsampling, the fusion with deep features after the third downsampling and fourth downsampling can further improve the network performance.

\subsection{Performance experiments}
\label{sec4_4}

In this section, we test the performance of the models, mainly including the accuracy on the validation set and the inference speed. To meet various application scenes, we provide three versions of StairNetV3, including StairNetV3-large, StairNetV3-medium, and StairNetV3-small, which can be obtained by adjusting the width factor of the network. To prevent the group convolution in the residual block from becoming depthwise separable convolution \cite{bib41}, which may reduce the feature expression ability of the small model. In StairNetV3, the width factor not only affects the number of channels in each layer of the network, but also the number of groups in each group convolution to ensure that the number of filters in each group remains unchanged. We use the precision, recall, and IOU at a confidence of 0.5, as well as PA and MPA as evaluation indicators, the results are shown in Table~\ref{tab2}.

\begin{table}[!htb]
  \caption{The results of StairNetV3 accuracy experiments.}
  \label{tab2}
  \centering
    \resizebox{0.85\textwidth}{!}{
    \begin{tabularx}{\textwidth}{>{\centering}ccccccc}
    \toprule
    \textbf{Model} & \textbf{Width factor} & \textbf{Precision ($ \% $)} & \textbf{Recall ($ \% $)} & \textbf{IOU ($ \% $)} & \textbf{PA ($ \% $)} & \textbf{MPA ($ \% $)} \\
    \midrule
    StairNetV3-large  & 1.0   & 77.3 & 81.8 & 65.9 & 96.3 & 81.3\\
    \midrule
    StairNetV3-medium & 0.75  & 75.5 & 81.8 & 64.6 & 96.0 & 80.9\\
    \midrule
    StairNetV3-small  & 0.5   & 74.1 & 81.0 & 63.1 & 95.9 & 80.7\\
    \bottomrule
    \end{tabularx}
    }
\end{table}

For the network inference speed test, we use a desktop platform with an i9 13900 CPU and an RTX 3090 GPU, and a mobile platform with Ryzen7 7735H CPU and RTX4060 laptop GPU for testing. To reflect the inference speed of the network under different workloads, we test the model inference time in two scenes: the batch size is set to 1, and the batch size is set to the maximum value that the GPU can handle. The former corresponds to online image or video stream applications and the latter corresponds to offline image or video stream applications. The detailed test results are shown in Table~\ref{tab3}.

\begin{table}[!htb]
  \caption{The results of StarNetV3 inference speed experiments.}
  \label{tab3}
  \centering
    \resizebox{\textwidth}{!}{
  \begin{tabular}{llll|cc|cc}
    \toprule
  & & & & \multicolumn{2}{c}{\textbf{Runtime (ms) Batch size=1}}  & \multicolumn{2}{c}{\textbf{Runtime (ms) Batch size=maximum}}  \\
\toprule
\textbf{Model}    & \textbf{Width-factor}   & \textbf{$\#$Params.}   & \textbf{GFlops} & Desktop-platform & Mobile-platform & Desktop-platform  & Mobile-platform \\
\midrule
\textbf{StairNetV3-large}    & 1.0   & 3.0 $ \times 10^7 $   & 62.3 & 27.1  & 80.2 & 15.5  & 62.2 \\
\midrule
\textbf{StairNetV3-medium}   & 0.75  & 1.7 $ \times 10^7 $   & 35.4 & 24.1  & 65.4 & 11.4  & 43.0 \\
\midrule
\textbf{StairNetV3-small}    & 0.5   & 7.8 $ \times 10^6 $   & 16.1 & 19.6  & 57.2 & 7.0   & 26.3 \\
    \bottomrule
  \end{tabular}
  }
\end{table}

\begin{figure}[!htb]
\includegraphics[width=\linewidth]{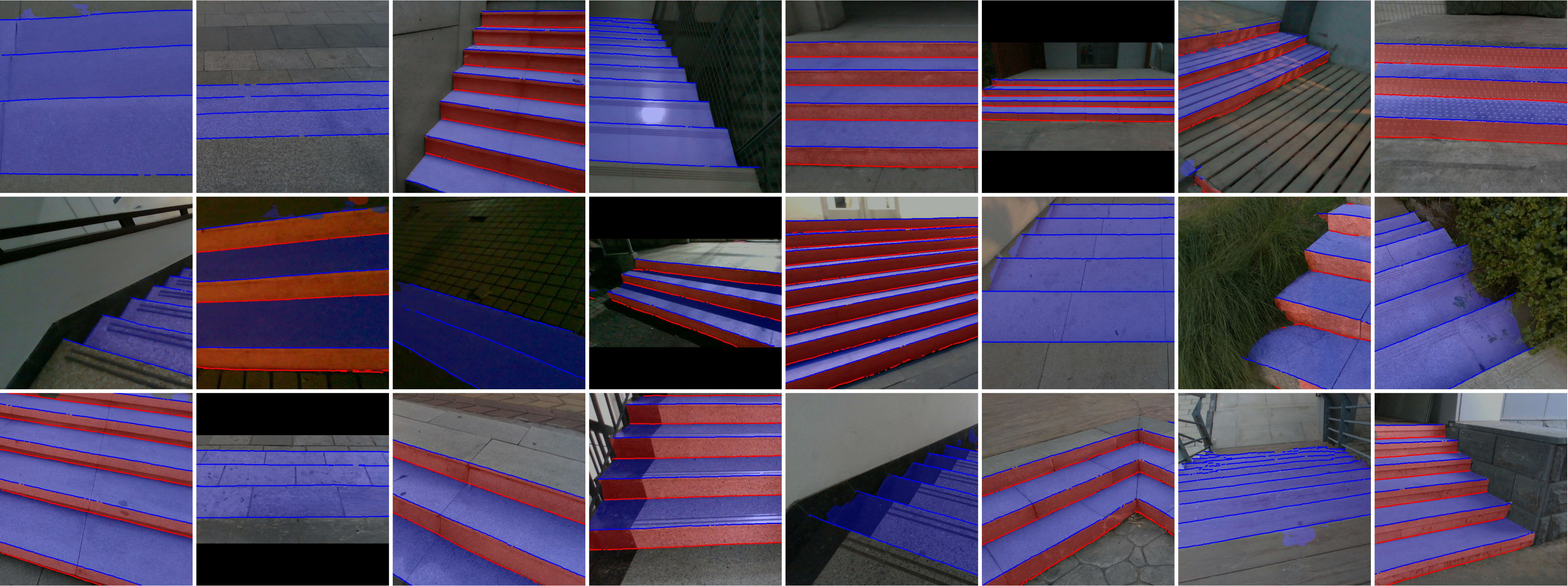}
\caption{Some visualization results of StairNetV3-large.}
\label{fig7}
\end{figure}

As we can see from Tables~\ref{tab2} and ~\ref{tab3}, as the model becomes lighter, the model accuracy gradually decreases, the speed gradually increases, and the performance fluctuation of the models is stable. When batch size is set to 1, the difference in inference time between the three models is not significant, allowing for flexible selection based on accuracy demands, real-time demands, and computational resource allocation. When batch size is set to the maximum value, there is a significant difference in the inference speed of the models. The large and medium models are suitable for tasks that prioritize accuracy, while the small model is suitable for tasks that prioritize speed. Some visualization results of StairNetV3-large are shown in Figure~\ref{fig7}, which intuitively demonstrates the adaptability of StairNetV3 to various stairs and shooting conditions. Particularly, even without depth images as input, StairNetV3 can still show stable performance in scenes with fuzzy visual clues, such as night and descending stairs.

\begin{figure}[!htb]
\includegraphics[width=\linewidth]{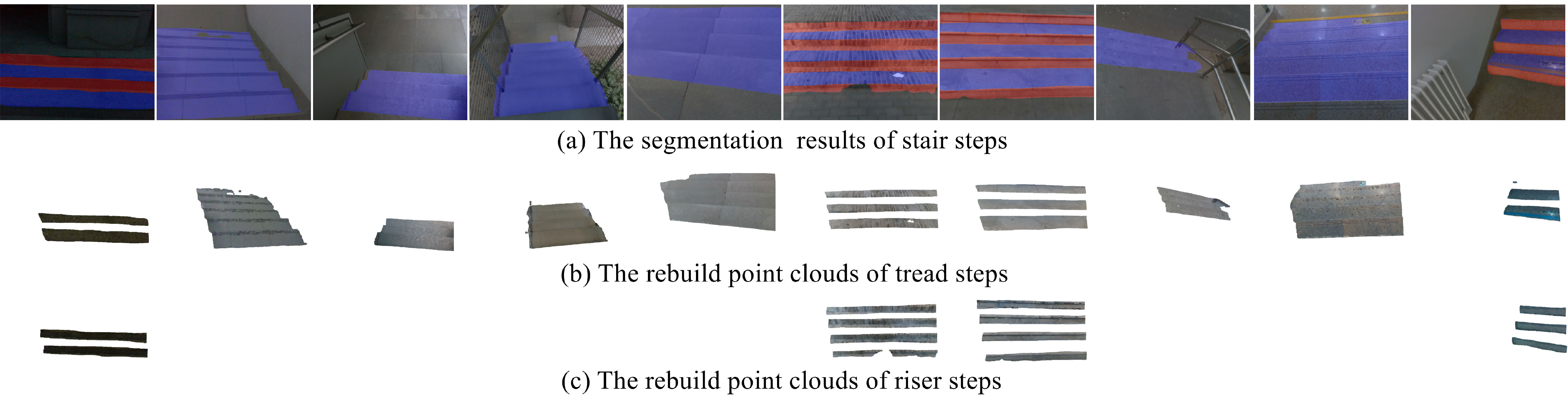}
\caption{Some visualization results of stair step surface segmentation and point cloud reconstruction.}
\label{fig8}
\end{figure}

After StairNetV3 obtains the segmentation results of the stair step surfaces, the stair modeling can be realized through point cloud reconstruction. Since the pixels of the image and the points of the point cloud correspond one-to-one, the PA and MPA results in Table~\ref{tab2} can be considered as the segmentation precision of the stair point cloud on the validation set. Because the images in the validation set are padded and scaled and the linear transformation relationships used to generate the depth images are not saved, these images cannot be truly used for point cloud reconstruction. We test point cloud reconstruction in the test set and because the test set does not have ground-truth labels, we only test the running speed, and the results are shown in Table~\ref{tab4}. Some visualization results of the stair step surface segmentation are shown in Figure~\ref{fig8}.

\begin{table}[!htb]
  \caption{Speed test results of point cloud reconstruction.}
  \label{tab4}
  \centering
    \resizebox{0.6\textwidth}{!}{
  \begin{tabular}{l|cc}
    \toprule
  & \multicolumn{2}{c}{\textbf{Runtime (ms)}} \\
\toprule
\textbf{Process}   & Desktop-platform & Mobile-platform\\
\midrule
\textbf{Post-processing of CNN outputs}    & 2.1   & 3.1 \\
\midrule
\textbf{Processing of depth image}   & 6.9  & 10.4 \\
\midrule
\textbf{Rebuilding point cloud}    & 3.5  & 5.3 \\
\midrule
\textbf{Whole process}    & 12.5  & 18.8 \\
    \bottomrule
  \end{tabular}
  }
\end{table}

As can be seen from Table~\ref{tab4}, the point cloud reconstruction algorithm has superior real-time performance on both desktop and mobile platforms. This is benefit from our approach of implementing point cloud segmentation by combining image segmentation with point cloud reconstruction, which avoids the slow speed and large data processing drawbacks of direct point cloud segmentation. From Figure~\ref{fig8}, it can be intuitively seen that our method can accurately segment point clouds belonging to the stair step surfaces in various scenes.

\subsection{Comparison experiments}
\label{sec4_5}

In this section, we compare StairNetV3 with some previous methods, including traditional image processing methods, methods combining traditional image processing and deep learning, and the deep learning method StairNetV1. We implement traditional image processing method using Gabor filters, Canny edge detection and Hough transform, and then we combine them with YOLOv5 \cite{bib24} as a combination of deep learning method. The outputs of all the methods in the comparison experiments are adjusted to be consistent with StairNetV3, and all the methods are carried out on the desktop platform described in section~\ref{sec4_4}. The test dataset is the RGB-D stair dataset, and the detailed experimental results are shown in Table~\ref{tab5}.

\begin{table}[!htb]
  \caption{Experimental results of comparison experiments.}
  \label{tab5}
  \centering
    \resizebox{.9\textwidth}{!}{
    \begin{tabularx}{\textwidth}{>{\centering}cccccc}
    \toprule
    \textbf{Method} & \textbf{Precision ($ \% $)} & \textbf{Recall ($ \% $)} & \textbf{IOU ($ \% $)} & \textbf{GFlops} & \textbf{Runtime (ms)} \\
    \midrule
    \makecell[c]{Gabor+Canny+Hough\\(Our implementation)}             & 18.9 & 19.8 & 10.7 & - & 4.0 \\
    \midrule
    \makecell[c]{YOLOv5+Gabor+Canny+Hough\\(Our implementation)} & 28.5 & 22.5 & 14.3 & 13.2 & 14.9\\
    \midrule
    StairNet 0.25 $ \times $  & 74.8 & 77.7 & 61.6 & 2.84 & 2.6\\
    \midrule
    StairNet 0.5  $ \times $  & 76.3 & 77.3 & 62.4 & 10.4 & 4.7\\
    \midrule
    StairNet 1 $ \times $  & 77.4 & 76.4 & 62.5 & 39.6 & 9.6\\
    \midrule
    StairNetV3-small (Ours)	& 74.1	& 81.0	& 63.1	& 16.1	& 7.0\\
    \midrule
    StairNetV3-medium (Ours)	& 75.5	& 81.8	& 64.6	& 35.4	& 11.4\\
    \midrule
    StairNetV3-large (Ours)	& 77.3	& 81.8	& 65.9	& 62.3	& 15.5\\
    \bottomrule
    \end{tabularx}
    }
\end{table}

It can be seen that deep learning methods have much higher accuracy than traditional methods. Because traditional methods  have heavy reliance on manually designed rules and the selection of thresholds, it is difficult to perform well in large dataset containing complex and variable scenes. While, deep learning methods extract features by learning, which can greatly improve adaptability to different scenes compared to traditional methods. Among the deep learning methods, in fact, the feature representation method with Gaussian kernels used by StairNetV3 is not conducive to improving accuracy. This is because the confidences of some positive samples are weakened, especially the positive samples in the middle of the stair lines have a confidence of only 0.6. For the feature representation method of StairNet, the confidences of positive samples are all 1, so there is a strong distinction between the foreground and background. Even under these circumstances, due to the awareness of depth information, StairNetV3 still has better performance than StairNet. Furthermore, StairNetV3-small with less parameters and faster speed still surpasses StairNet 1 $ \times $ in accuracy, which proves the effectiveness of our method.

\begin{figure}[!htb]
\includegraphics[width=\linewidth]{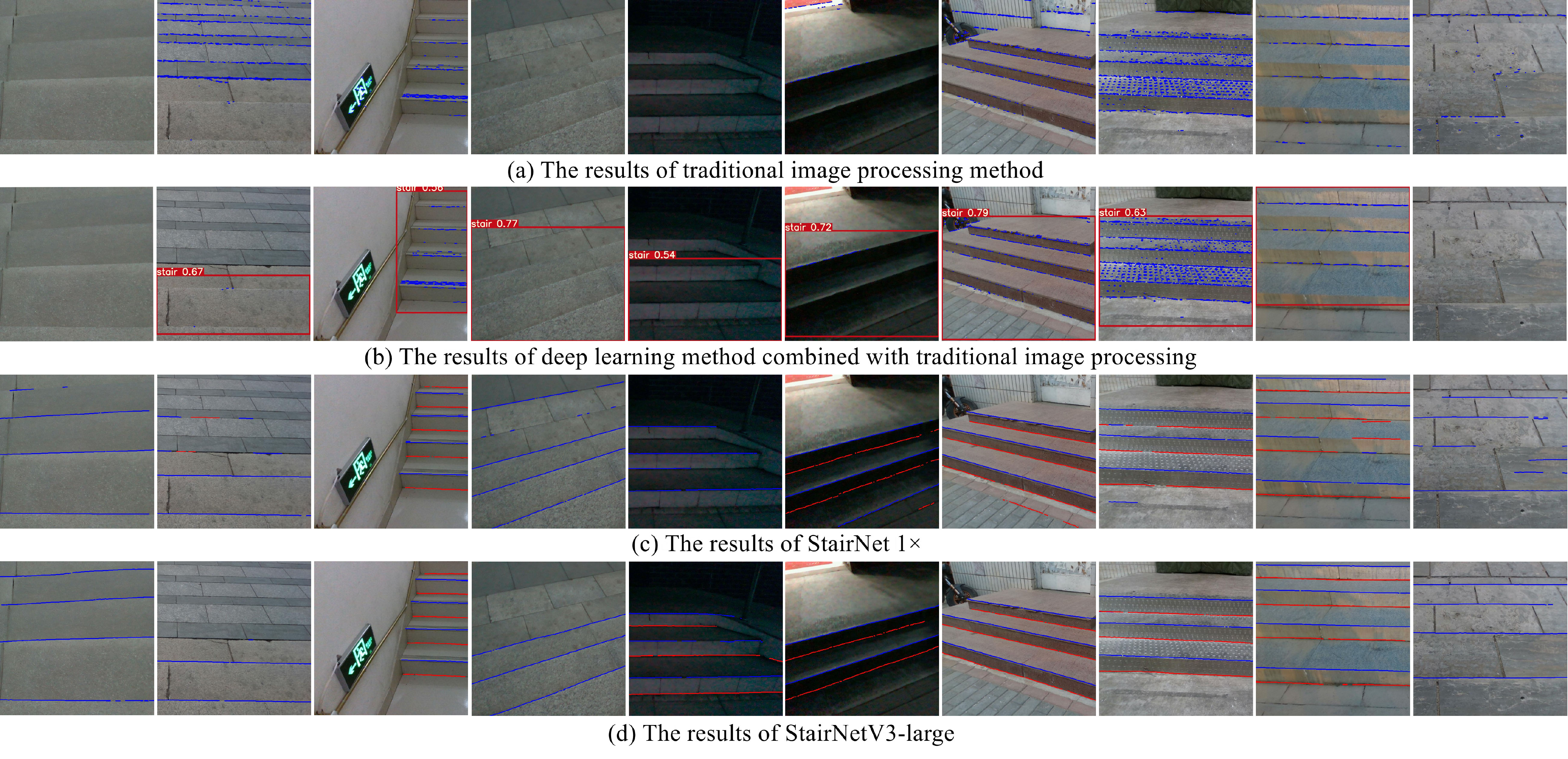}
\caption{Some visualization results of comparison experiments.}
\label{fig9}
\end{figure}

Some visualization results of the comparison experiments are shown in Figure~\ref{fig9}. It can be seen that traditional method has a certain degree of false detection and missed detection in different scenes due to their sensitivity to manually designed rules and thresholds. The deep learning method combined with traditional image processing can eliminate false detection to some extent, but the missed detection is still serious. The deep learning methods show great adaptability to various complex scenes, and StairNetV3 has better performance in environments with fuzzy visual cues compared to StairNet, such as night and descending stairs. In addition, when there is ground with similar texture to stairs, StairNetV3 has less false detection compared to StairNet.

\section{Conclusion}
\label{sec5}

We propose a depth-aware stair modeling convolutional neural network that can simultaneously extract stair lines and stair step surfaces under monocular vision, overcoming the poor performance in environments with fuzzy visual cues, such as night and descending stairs. To learn more abundant stair line information, including stair line endpoints, we propose a new stair line feature representation method with Gaussian kernels. To achieve fast extraction of stair step surfaces, we propose a point cloud reconstruction method to segment the point clouds of stair step surfaces in an image. Experiments conducted on the RGB-D stair dataset demonstrate the effectiveness of our method. In future work, we will study the estimation of stair geometric features in the world coordinate system to further explore the potential of monocular vision methods.

\bibliography{sample}

\end{document}